\documentclass[
]{ceurart}

\usepackage{listings}
\usepackage{graphicx}  
\begin{document}

\copyrightyear{2025}
\copyrightclause{Copyright for this paper by its authors.
  Use permitted under Creative Commons License Attribution 4.0
  International (CC BY 4.0).}

\conference{ProActLLM Proactive Conversational Information Seeking with Large Language Models, November 14, 2025, Coex, Seoul, South Korea (co-located with CIKM 2025)}

\title{ConCISE: A Reference-Free Conciseness Evaluation Metric for LLM-Generated Answers}

\tnotemark[1]
\tnotetext[1]{This paper presents ConCISE, a novel metric for evaluating the conciseness of LLM-generated answers without requiring reference texts.}

\author[1]{Seyed Mohssen Ghafari}[%
email=seyedmohssen.ghafari@cba.com.au,
]

\cortext[1]{Corresponding author.}

\cormark[1]

\author[1]{Ronny Kol}[%
email=ronny.kol@cba.com.au,
]

\author[1]{Juan C. Quiroz}[%
email=juan.quirozaguilera@cba.com.au,
]

\author[1]{Nella Luan}[%
email=nella.luan@cba.com.au,
]

\author[1]{Monika Patial}[%
email=Monika.Patial@cba.com.au,
]

\author[1]{Chanaka Rupasinghe}[%
email=Chanaka.Rupasinghe@cba.com.au,
]

\author[1]{Herman Wandabwa}[%
email=Herman.Wandabwa@cba.com.au,
]

\author[1]{Luiz Pizzato}[%
email=Luiz.Pizzato1@cba.com.au,
]
\cormark[1]
\address[1]{Commonwealth Bank of Australia, Sydney, Australia}

\begin{abstract}
  Large language models (LLMs) frequently generate responses that are lengthy and verbose, filled with redundant or unnecessary details. This diminishes clarity and user satisfaction, and it increases costs for model developers, especially with well-known proprietary models that charge based on the number of output tokens. In this paper, we introduce a novel reference-free metric for evaluating the conciseness of responses generated by LLMs. Our method quantifies non-essential content without relying on gold standard references and calculates the average of three calculations: i) a compression ratio between the original response and an LLM abstractive summary; ii) a compression ratio between the original response and an LLM extractive summary; and iii) word-removal compression, where an LLM removes as many non-essential words as possible from the response while preserving its meaning, with the number of tokens removed indicating the conciseness score. Experimental results demonstrate that our proposed metric identifies redundancy in LLM outputs, offering a practical tool for automated evaluation of response brevity in conversational AI systems without the need for ground truth human annotations.
\end{abstract}


\maketitle

\section{Introduction}

As large language models (LLMs) \citep{OpenAI23, Djeddal24} are increasingly used to answer questions and engage in dialogue, the quality of their responses becomes critical. In many applications, brief and clear answers are preferred \citep{Nayab24}. However, LLMs often produce overly verbose, long-winded responses containing redundant or irrelevant information \citep{Nayab24}. A response that is thorough but lengthy may overwhelm users, while one that is brief but lacks detail may fail to meet their needs. Thus, conciseness–providing the shortest answer that still covers the necessary content–is a desirable property of LLM outputs. 
Conciseness is rarely directly measured by existing evaluation metrics. Traditional metrics for text generation (e.g. BLEU or ROUGE) depend on reference texts and focus on lexical overlap or content coverage, which do not capture verbosity \citep{Papineni02}. Recent work has explored reference-free metrics for other quality aspects \citep{Papineni02}. For example, the RAGAS framework introduces reference-free metrics for retrieval-augmented question answering, allowing automated evaluation without ground-truth answers. Inspired by such approaches, we seek a metric that quantifies conciseness by detecting non-essential content in an answer. We leverage LLM capabilities to simulate human judgments of brevity in a reference-free manner.

\begin{figure}[h]
    \centering
    \includegraphics[width=0.95\textwidth]{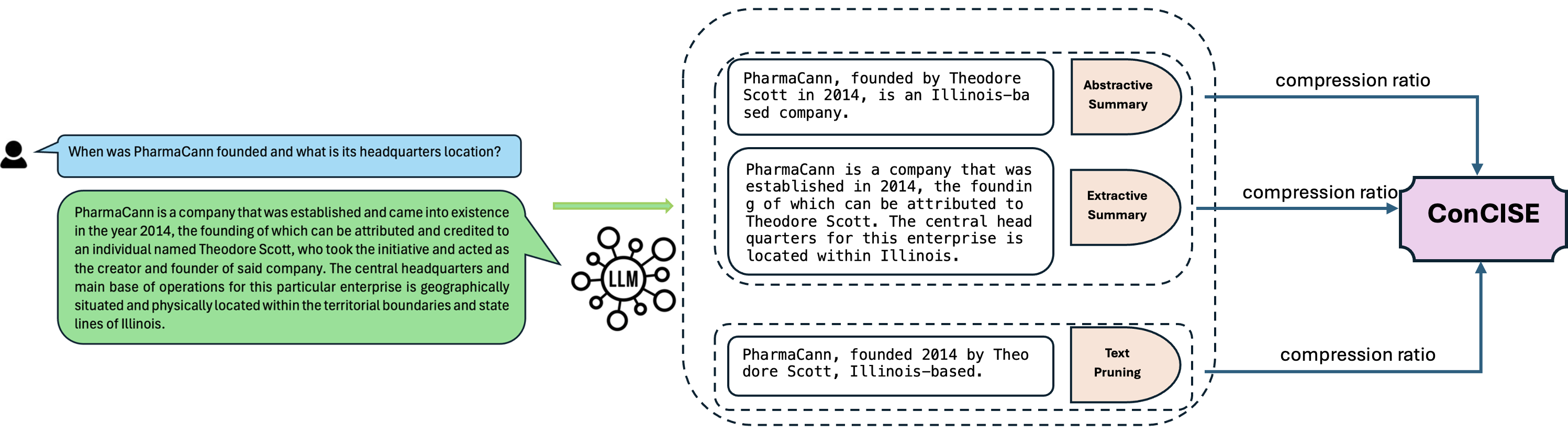}
    \caption{ConCISE Architecture}
    \label{fig:yourlabel}
\end{figure}

We propose ConCISE, a new conciseness metric that operates without gold-standard answers. Our approach quantifies non-essential content without relying on gold standard references and calculates the average of three calculations: i) a compression ratio between the original response and an LLM abstractive summary; ii) a compression ratio between the original response and an LLM extractive summary; and iii) word-removal compression, where an LLM removes as many non-essential words as possible from the response while preserving its meaning, with the number of tokens removed indicating the conciseness score. We apply ConCISE to the WikiEval dataset (a set of Wikipedia-based questions) and verify that our approach effectively measures the conciseness of an LLM-generated response.  
The contributions of this paper are as follows: 1) we propose a novel reference-free metric for evaluating the conciseness of responses generated by LLMs; 2) we conducted tests to demonstrate the effectiveness of the new metric and its level of alignment with human judgment; 3) to the best of our knowledge, this metric is one of the first evaluation mechanisms that can assess an LLM’s output based on its length without requiring any gold standard reference answers.

\section{Related Work}

\subsection{General Evaluation Metrics}

Classical reference-based evaluation metrics like BLEU and ROUGE evaluate generated text by lexical overlap with gold standard references \citep{Papineni02}. More advanced metrics such as BERTScore and BLEURT focus on semantic similarity, but rely on reference texts \citep{Zhang20}. As preparing human-annotated references can be costly and time-consuming \citep{Chen21}, several unsupervised approaches have emerged. Chen et al. \citep{Chen21} proposed a reference-free redundancy and relevance metric to evaluate a given summary. Moreover, RAGAS evaluates Retrieval-Augmented Generation (RAG) systems using factuality and relevance scores independent of gold standard answers \citep{Papineni02}. Recent work also leverages LLMs as evaluators (LLM-as-a-judge), achieving human-level evaluation reliability in specific tasks like code generation and summarization \citep{Fu23, Ji23, Zhang24, Naik25}.

\subsection{Verbosity and Length-Sensitive Metrics}

Recent research highlighted verbosity as a critical quality factor. Nayab et al. introduced concise reasoning metrics for reducing verbosity in LLM reasoning chain-of-thought (CoT) \citep{Nayab24}. Other studies have shown that verbosity negatively impacts translation and question-answering evaluations, often biasing LLM-based evaluators towards overly long outputs \citep{Briakou24, Saito23, Hu24}. Methods explicitly addressing verbosity include normalization techniques (AdapAlpaca) and length-controlled generation to mitigate such biases \citep{Hu24, Butcher24}. 

\subsection{Summarization and Compression Metrics}

While summarization inherently involves conciseness, traditional metrics overlook verbosity explicitly. Recent studies introduced conciseness as an evaluation dimension in summarization, emphasizing semantic compression and brevity \citep{Chen21, Stahlberg22}. Techniques for input prompt compression like LLMLingua and methods evaluating information density have indirectly advanced conciseness assessment \citep{Stahlberg22, Jiang23, Li25}.

\subsection{Comparison to Our Proposed Metric (ConCISE)}

Our metric, ConCISE, uniquely quantifies verbosity directly by leveraging LLM-based summarization and word-removal methods, explicitly penalizing redundant information without gold standard references. Unlike previous LLM-based evaluators susceptible to length bias, ConCISE systematically evaluates brevity aligned with human judgment, bridging existing evaluation gaps highlighted by prior research.

\section{Experimental Design}

We consider an ordinary solution developed based on LLMs. When a given query q arrives, the system will provide an answer a(q) to that query with the help of an LLM. Our main priority is to design an evaluation metric that is fully self-contained and reference-free, as in real-world scenarios we usually do not have access to human-annotated data \citep{Papineni02}. We focus on assessing an LLM’s response in terms of Conciseness. Conciseness refers to the ability of an LLM to be short and generate the least number of words without sacrificing accuracy. Indeed, LLMs may generate responses that are lengthy and verbose, filled with redundant or unnecessary details. This not only diminishes clarity and user satisfaction \citep{Nayab24} but also increases costs for customers, especially with well-known proprietary models that charge based on the number of output tokens. In our experiment, we have used well-known LLMs, e.g., GPT-4o, Claude-4, and Gemini-2.0.

\subsection{Conciseness}

We define Conciseness as: a LLM’s answer a(q) is concise if you cannot remove any words from a(q) without harming its meaning and sacrificing its accuracy. To estimate conciseness, we propose two mechanisms: i) summarization-based compressions, where we use an LLM to provide an extractive Se(a(q)) and an abstractive summary Sa(a(q)) of the answer. The aim is to see if we can summarize the LLM’s answer without sacrificing its accuracy; ii) word-removal compression, where we use an LLM to remove as many non-essential words as possible from the response while preserving its meaning, Wr(a(q)). The number of words removed indicates the level of conciseness of the answer. The prompt that we used is as follows:

\vspace{1em}

\begin{minipage}{0.8\textwidth}
\itshape
Given a question-answer pair, generate three versions of the answer using the following techniques:
\begin{enumerate}
\item \textbf{Abstractive Summary}: Create a paraphrased summary that captures the main ideas using new phrasing.
\item \textbf{Extractive Summary}: Select and present the most relevant sentences directly from the original text.
\item \textbf{Pruned Text}: Produce a minimalist version of the original text by removing all non-essential words while preserving the core meaning.
\end{enumerate}
\textbf{Note}: Prioritize maximum conciseness while maintaining semantic integrity.

\texttt{question: [question], answer: [answer]}
\end{minipage}

\vspace{1em}

, where [question] and [answer] represent q and a(q), respectively. Hence, for each answer, the LLM will provide Se(a(q)), Sa(a(q)), and Wr(a(q)). We then ask an LLM to judge if these values preserved the same meaning and all the important entities as a(q). This step can be done using the following prompt:

\begin{minipage}{0.8\textwidth}
\itshape
Given an original answer and three derivative texts (extractive summary, abstractive summary, and pruned version), evaluate each text against these criteria:
\begin{enumerate}
\item \textbf{Semantic Equivalence}: Verify if the meaning remains consistent with the original answer
\item \textbf{Named Entity Preservation}: Confirm that all original named entities (dates, locations, etc.) are retained
\end{enumerate}

For each text, provide a binary assessment (Yes/No) based on:
\begin{itemize}
\item Maintenance of core meaning
\item Complete preservation of named entities
\end{itemize}

\textbf{Original Answer}: \texttt{[answer]}

\textit{Extractive Summary, Abstractive Summary, Pruned Text}
\end{minipage}

\vspace{1em}

The final Conciseness, ConCISE, can be computed as:

\[
\text{ConCISE} = \frac{1}{n}\left[\left(1-\frac{|A|-|AS|}{|A|}\right)+\left(1-\frac{|A|-|ES|}{|A|}\right)+\left(1-\frac{|A|-|RW|}{|A|}\right)\right]
\]

\vspace{1em}

, where |A| represents the word length of the answer, |AS| is the word length difference between the answer and the abstractive summary statement, |ES| is the word length difference between the answer and the extractive summary statement, and |RW| is the length difference between the answer and the copy of the answer that does not contain unnecessary words. In this formula, n is 3. In cases where |AS|, |ES|, or |RW| are negative values, or these statements are longer than the original answer, we consider their values as zero.

\subsection{Dataset}

For our experiments, we utilized the WikiEval dataset\footnote{\url{https://huggingface.co/datasets/explodinggradients/WikiEval}}, a human-annotated benchmark specifically designed to assess the quality of answers generated by language models. This dataset includes question-context-answer triples sourced from 50 recently updated Wikipedia pages. Using the WikiEval dataset, we prompted GPT-4o to take each answer a(q) and generate a verbose version that retains the same facts but includes filler and repetition. The prompt used was as follows:  

\vspace{1em}

\begin{minipage}{0.8\textwidth}
\itshape
You will be given an answer to a question. Rewrite the answer to be more verbose by adding redundancy and extra explanations while preserving all key facts, entities, and the original meaning exactly. Instructions:

- Do not omit any key points, entities, or facts.

- Add redundant phrases, rephrase the same information multiple times, or include filler to increase verbosity.

- Do NOT add new facts or incorrect information.

- Ensure the rewritten answer is notably longer
\end{minipage}

\vspace{1em}

To obtain human judgments about conciseness, we asked three human annotators to provide their feedback as follows:

\vspace{1em}

\begin{minipage}{0.95\textwidth}
1. Likert Scale Ratings: Rate each answer's conciseness on a 5-point Likert scale (e.g. 5 = very concise, 1 = very verbose). Provide the original question alongside the answer so raters can judge if the answer "only conveys the necessary".

2. Pairwise Comparison (Ranking): Compare pairs of answers and judge which is more concise for answering the given question.
\end{minipage}

\subsection{Baseline}
Following the approach of Shahul Es et al. \citep{Papineni02}, we evaluated ConCISE against two baseline methods. The first baseline is GPT Score, which uses an LLM to rate answers on a scale of 0 to 10 based on specific prompts:

\vspace{1em}

\begin{minipage}{0.8\textwidth}
\itshape
Conciseness measures how efficiently an answer conveys its intended information. A concise answer avoids unnecessary elaboration or redundancy, while fully preserving all core facts. Given an answer, assign a score for conciseness in the range 0–10.

\textbf{answer:} [answer]
\end{minipage}

\vspace{1em}

Our second baseline is GPT Ranking, which we ask an LLM to select the preferred answer with the help of the following prompt:

\vspace{1em}

\begin{minipage}{0.8\textwidth}
\itshape
Conciseness measures how efficiently an answer conveys its intended information. A concise answer avoids unnecessary elaboration or redundancy, while fully preserving all core facts. Given a question and two answers, choose the more concise one.

\textbf{question:} [question]

\textbf{answer 1:} [answer 1], \textbf{answer 2:} [answer 2]
\end{minipage}

\vspace{1em}

Finally, to ensure robust evaluation of ConCISE, we employed multiple LLM models (GPT-4o, Claude-4-Sonnet, Gemini-2.0-Flash, Mistral-Large-2) as judges to avoid model-specific bias and comprehensively assess the metric's performance across different downstream LLM architectures.

\section{Results}

We computed Spearman's rank correlation ($r_s$) between the ConCISE scores and the human Likert Scale Ratings across all answers. A high Spearman $r_s$ (closer to 1) would mean ConCISE effectively ranks answers by conciseness similarly to humans. We also computed Kendall's Tau ($\tau$) as an alternative rank correlation metric, which is more sensitive to pairwise order flips. These correlation coefficients tell us quantitatively how well ConCISE approximates human judgment. Table 1 demonstrates the experimental results and compares ConCISE with the baseline metrics. In this table, $r_s$ and $\tau$ represent the correlation coefficients for Spearman and Kendall respectively, while $\rho_s$ and $\rho_k$ represent the corresponding p-values for testing the statistical significance.

Next, we measured the alignment between ConCISE and other baseline metrics with human judgments on pairwise comparisons. Specifically, we examined how often these metrics agreed with human annotators when determining which answer was more concise between two options (Accuracy). This analysis helped us understand the reliability of ConCISE and baseline metrics in matching human preferences for conciseness.

\begin{equation}
\text{Accuracy} = \frac{\text{Number of Matches}}{\text{Total Number of Comparisons}} \times 100
\label{eq:accuracy}
\end{equation}

\begin{table}[h]
\caption{Spearman ($r_s$) and Kendall ($\tau$) correlations with human annotations and different evaluation metrics }
\centering
\begin{tabular}{lcccc}
\toprule
Metric & $r_s$ & $\rho_s$ & $\tau$ & $\rho_k$ \\
\midrule
ConCISE\-GPT-4o & 0.628 & $< 0.001$ & 0.523 & $< 0.001$ \\
ConCISE\-Claude-4\-Sonnet & 0.537 & $< 0.001$ & 0.436 & $< 0.001$ \\
ConCISE\-Gemini-2.0-Flash & 0.518 & $< 0.001$ & 0.422 & $< 0.001$ \\
ConCISE\-Mistral-Large-2 & 0.473 & 0.0017 & 0.376 & 0.002 \\
GPT Score & $-0.108$ & 0.474 & $-0.087$ & 0.490 \\
\bottomrule
\end{tabular}
\label{tab:performance}
\end{table}

\begin{table}[h]
\centering
\caption{Human Alignment of ConCISE vs GPT Ranking}
\begin{tabular}{lc}
\toprule
\textbf{Metric} & \textbf{Accuracy (\%)} \\
\midrule
ConCISE\-Claude-4\-Sonnet & \textbf{94} \\
ConCISE\-Gemini-2.0-Flash & \textbf{94} \\
ConCISE\-Mistral-Large-2 & \textbf{94} \\
ConCISE\-GPT-4o & 90 \\
GPT Ranking & 39 \\
\bottomrule
\end{tabular}
\label{tab:accuracy}
\end{table}

\section{Discussion}

Our results suggest that our proposed metric, ConCISE, shows promise in capturing human judgments of conciseness in LLM-generated responses. The best version of ConCISE achieved a Spearman's rank correlation ($r_s$) of 0.628 and a Kendall's $\tau$ of 0.523 with human annotations, both statistically significant ($\rho_s$ and $\rho_k < 0.001$). In contrast, the GPT Score baseline showed weak correlations with human ratings ($r_s = -0.108$, $\tau = -0.087$, $\rho_s$ and $\rho_k > 0.4$), suggesting that direct numeric scoring via prompts may have limitations as a proxy for human conciseness judgments in this context.

In terms of pairwise comparisons—where systems are evaluated based on how often they agree with human preferences when choosing the more concise of two answers— the best version of ConCISE again demonstrated superior performance, aligning with human decisions 94\% of the time (Table 2), substantially outperforming the GPT Ranking method.

These initial findings suggest that in our experimental setup—using this specific dataset, evaluator set, and LLM-judge prompting approach—general-purpose LLM judgments and pointwise scoring showed limited effectiveness for conciseness evaluation. ConCISE offers a straightforward, reference-free framework requiring only a single LLM call, which shows promising alignment with human perceptions of effective communication. This simplicity makes it potentially valuable for practical evaluation of conversational AI systems, though broader validation across diverse datasets, prompting strategies, and evaluation settings would be needed to establish its generalizability and robustness.

One major limitation of ConCISE is the context-dependent nature of conciseness. The definition of non-essential content varies across domains—verbose details like regulatory disclosures in finance or explanatory elaborations in education may appear redundant but remain critical for accuracy and functional relevance. Future work can explore more advanced domain-adaptive models and human feedback loops to improve the robustness of conciseness evaluations across diverse applications. Furthermore, while ConCISE's unified prompt approach is cost-effective and practical for real-world deployment, using separate prompts for each compression technique could reduce potential bias where the model's performance on one technique influences the others within the same generation cycle. Future work could address this by implementing and comparing separate prompts for each compression technique to isolate potential cross-technique bias.

\section{Conclusion}

In this paper, we have proposed a novel reference-free metric for evaluating the conciseness of LLM-generated responses. The proposed metric does not rely on gold standard references. Experimental results demonstrate the effectiveness of this metric in identifying redundancy in LLMs' outputs, offering a practical tool for automated evaluation of response brevity in conversational AI systems without the need for ground truth human annotations. 

\section{Generative AI Declaration}

We used LLMs, such as GPT-4o for creating verbose answers in our dataset. We have also used LLMs as part of our main evaluation metric for creating abstractive and extractive summaries, as well as for generating pruning versions of answers. Next, we used LLMs in the experiments for designing our baseline metrics. Finally, we used LLMs for light editing of our text (e.g., automated grammar checks, and word autocorrect).

\bibliography{sample-ceur}

\end{document}